\documentclass[preprint,12pt,authoryear]{elsarticle}

\usepackage{amssymb}
\usepackage{amsmath}
\usepackage{CJK}
\usepackage{multirow}
\usepackage{subfig}
\usepackage{appendix}
\usepackage[colorlinks,
linkcolor=blue,
anchorcolor=blue,
citecolor=red
]{hyperref}
\newcommand{\tabincell}[2]{\begin{tabular}{@{}#1@{}}#2\end{tabular}}

\journal{neurocomputing}

\begin{document}

\begin{frontmatter}



\title{Select and Calibrate the Low-confidence: Dual-Channel Consistency based Graph Convolutional Networks}

\author[label1]{Shuhao Shi}
\author[label1]{Jian Chen}
\author[label1]{Kai Qiao}
\author[label1]{Shuai Yang}
\author[label1]{Linyuan Wang}
\author[label1]{Bin Yan*}
\address[label1]{organization={PLA strategy support force information engineering university, Henan Key Laboratory of Imaging and Intelligence Processing},
             city={Zhengzhou},
             country={China}}

\begin{abstract}
The Graph Convolutional Networks (GCNs) have achieved excellent results in node classification tasks, but the model's performance at low label rates is still unsatisfactory. Previous studies in Semi-Supervised Learning (SSL) for graph have focused on using network predictions to generate soft pseudo-labels or instructing message propagation, which inevitably contains the incorrect prediction due to the over-confident in the predictions. Our proposed Dual-Channel Consistency based Graph Convolutional Networks (DCC-GCN) uses dual-channel to extract embeddings from node features and topological structures, and then achieves reliable low-confidence and high-confidence samples selection based on dual-channel consistency. We further confirmed that the low-confidence samples obtained based on dual-channel consistency were low in accuracy, constraining the model's performance. Unlike previous studies ignoring low-confidence samples, we calibrate the feature embeddings of the low-confidence samples by using the neighborhood's high-confidence samples. Our experiments have shown that the DCC-GCN can more accurately distinguish between low-confidence and high-confidence samples, and can also significantly improve the accuracy of low-confidence samples. We conducted extensive experiments on the benchmark datasets and demonstrated that DCC-GCN is significantly better than state-of-the-art baselines at different label rates.

\end{abstract}



\begin{keyword}
Semi-supervised \sep Graph Convolutional network \sep Dual-channel consistency \sep Low-confidence samples calibration


\end{keyword}

\end{frontmatter}



\section{Introduction}
\label{1-Introduction}
In recent years, Graph Convolutional Networks (GCNs) have achieved remarkable success in a variety of graph-based networks for tasks ranging from traffic prediction \cite{article32,article33,article34} and recommender systems \cite{article28,article35,article36} to biochemistry \cite{article37,article38}. Despite their effectiveness and popularity, the training of GCNs usually requires a significant amount of labeled data to achieve satisfactory performance. However, obtaining these labels may be time-consuming, laborious and expensive. A great deal of research has turned to semi-supervised learning (SSL) methods to address the problem of few labeled samples. SSL enables co-learning of models from labeled and unlabeled data, resulting in improved precision of decision worlds and robustness of models.
\par
\begin{figure}[h]
 \centering
 \subfloat[]{
 \begin{minipage}[h]{0.85\linewidth}
 \label{fig:1a}
 \centering
 \includegraphics[width=0.85\linewidth]{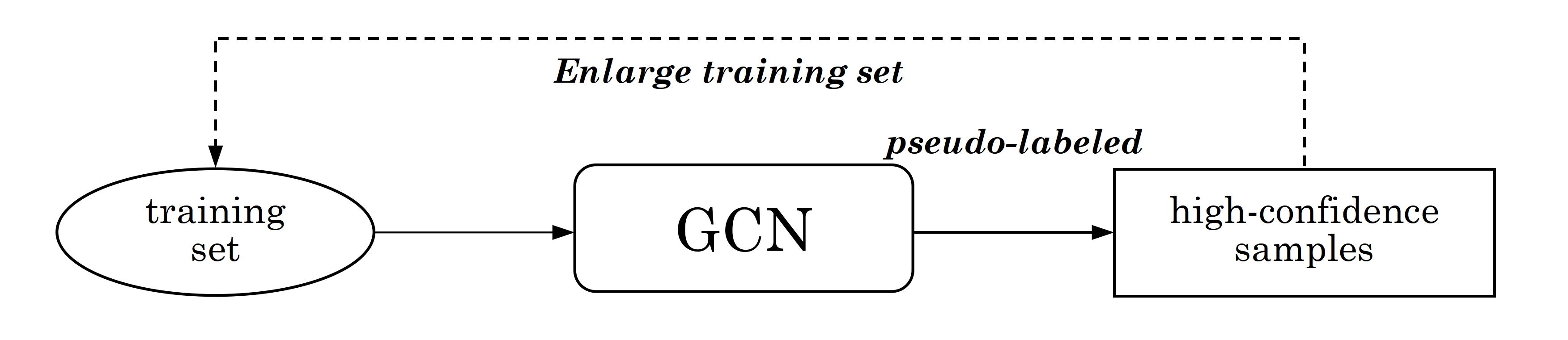}
 \end{minipage}%
 }%

 \subfloat[]{
 \begin{minipage}[h]{0.85\linewidth}
  \label{fig:1b}
 \centering
 \includegraphics[width=0.85\linewidth]{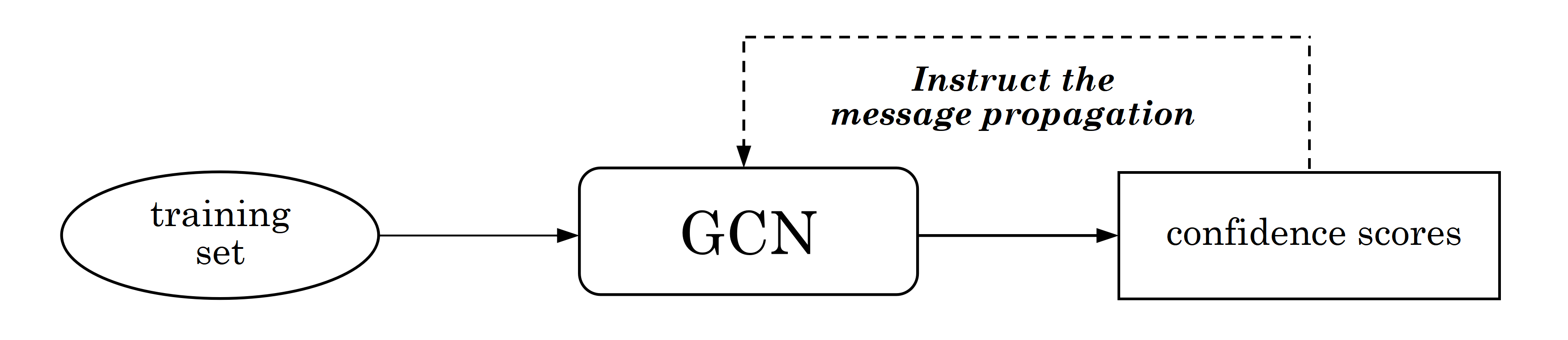}
 \end{minipage}
 }%

 \subfloat[]{
 \begin{minipage}[h]{0.85\linewidth}
  \label{fig:1c}
 \centering
 \includegraphics[width=0.85\linewidth]{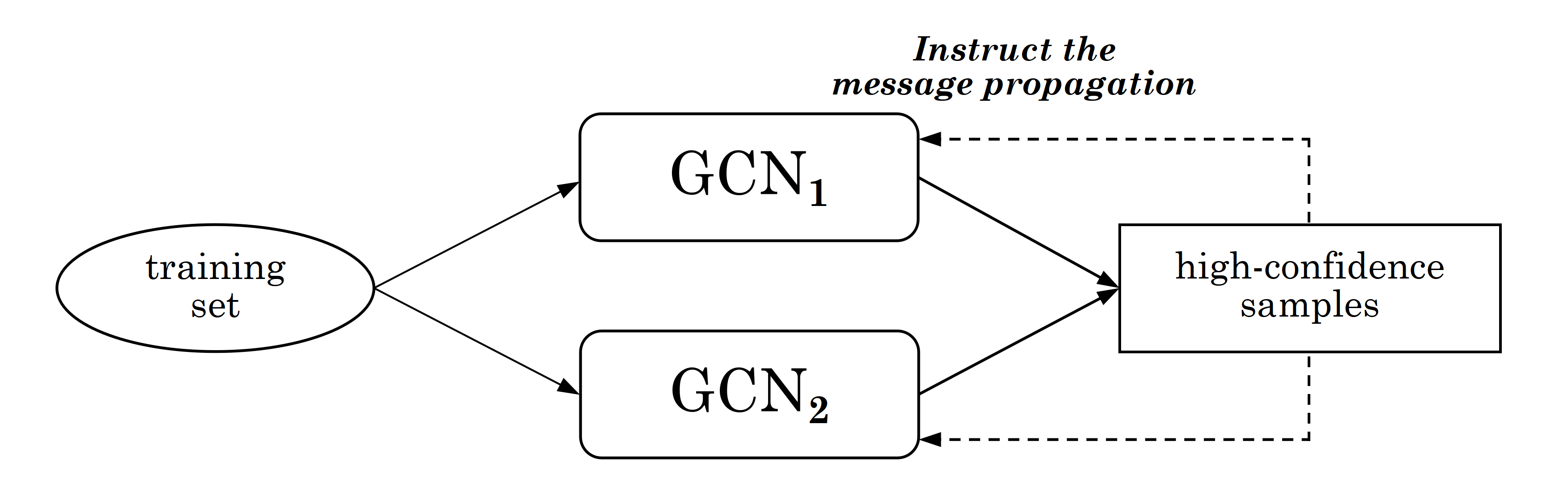}
 \end{minipage}
}%
\caption{The comparison between previous Graph-based SSL methods and our proposed DCC-GCN.}
\label{fig:1}
\centering
\end{figure}

 In GCNs, as illustrated in Figure \ref{fig:1a}, some pseudo-labeling methods \cite{article5,article17} have been proposed to obtain labels of unlabeled samples. In addition to pseudo-label learning methods, as illustrated in Figure \ref{fig:1b}, some recent work employs the output of the model to instruct the message propagation. Several studies \cite{article4,article7,article27} use the given labels and the estimated labels to optimize the topology graph, instructing the message propagation of the GCN model. ConfGCN \cite{article3} uses the estimated confidence to determine the effect of one node on another during neighborhood aggregation, changing the topological graph implicitly.

\par Although graph-based SSL methods have achieved noticeable progress in recent years, extensive researches \cite{article1,article18,article30} have revealed that neural network models are over-confident in their predictions. \cite{article1,article2,article21} have revealed that regardless of the accuracy of the outputs generated by the models, the confidence of the outputs gradually increases with training, resulting in the difficulty of selecting truly high-confidence samples as pseudo labels or correctly instructing GCN model message propagation. This seems to be a common problem in selecting truly high-confidence samples, which gives rise to a fundamental question: \itshape How to select high-confidence and low-confidence samples accurately? \upshape Existing methods tend to utilize high-confidence samples, yet seldom fully explore the use of low-confidence samples, which rises another critical question: \itshape How to calibrate low-confidence samples accurately? \upshape

\par In this paper, we introduce a framework called Dual-Channel Consistency based Graph Convolutional Networks (DCC-GCN) to select and calibrate low-confidence samples based on dual-channel consistency for GCN. As the first contribution of this study, we proved that classification errors are often where the predictions of different classifiers are inconsistent. Based on the above findings, as illustrated in Figure \ref{fig:1c}, we present a novel method to select low-confidence and high-confidence samples accurately based on inconsistent results of two different models. Previous methods only use a single topology graph for node aggregation, which fails to fully utilize the richness of features. Therefore, we trained two different classifiers using topology and feature graphs to select low-confidence samples. As the second contribution of this study, we confirm that the low-confidence samples obtained based on dual-channel consistency severely constrain the model's performance. In contrast to existing methods that focus on utilizing high-confidence samples, we improve the model's performance by calibrating for low-confidence samples using the neighborhood's high-confidence samples. In summary, the contributions of this paper are as follows:

\par 1. We propose an efficient graph-based SSL method called DCC-GCN, which could identify low-confidence and high-confidence samples more reliably based on the consistency of the dual-channel models. Our introduced dual-channel structure in SSL uses topology and feature graphs simultaneously, extracting more relevant information from node features and topology.

\par 2. We confirmed that the low-confidence samples limit the model's performance and improve the model's performance by calibrating for low-confidence samples using the neighborhood’s high-confidence samples.

\par 3. Our extensive experiments on a range of benchmark data sets clearly show that DCC-GCN can significantly improve the accuracy of low-confidence samples. Under different label rate settings, DCC-GCN consistently outperforms the most advanced graph-based SSL algorithm.

\par The rest of the paper is organized as follows. We review related work in Section \ref{2-Related works}, and develop DCC-GCN in Section \ref{3-Methods}. Then, We report experimental results on eight famous common datasets in Section \ref{4-Experiment}. In Section \ref{5-Discussion} we experimentally investigate the capability of DCC-GCN in calibrating low-confidence samples and aggregating features and topology. Finally, We conclude the paper in Section \ref{6-Conclusion}.

\section{Related works}
\label{2-Related works}
\subsection{Graph Convolutional Network}
\label{sec:2-1}
GCN \cite{article8} defines the convolution using a simple linear function of the graph Laplacian on semi-supervised classification on graphs, but this limits its capability at aggregating the information from nodes with similar features. To combat the shortcoming of GCN, GAT \cite{article11} introduces the attention mechanism in message propagation. More recently, task-specific GCNs were proposed in different fields such as \cite{article28,article32,article33,article34,article35,article36}.
\subsection{Graph-based Semi-supervised Learning}
\label{sec:2-2}
SSL on graphs is aim at classifying nodes in a graph, where labels are available only for a small subset of nodes. Conventionally, pseudo-label learning approaches continuously adopt model predictions to expand the labeled training set \cite{article5,article17,article29}. Recently, more attention has been attracted to instructing model training based on output results \cite{article3,article4,article7,article16,article27}. Typically, AGNN \cite{article27} builds a new aggregation matrix based on the learned label distribution, which updates the topology using the output. ConfGCN \cite{article3} uses the confidence of sample estimates to determine the influence of one node on another during neighborhood aggregation, changing the weights of the edges in the topology. Furthermore, E-GCN \cite{article4} utilizes both given and estimated labels to optimize the topology.
\par Despite the noticeable achievements of these graph-based semi-supervised methods in recent years, the main concern in SSL that selects more reliable high-confidence samples has not been well addressed. Selecting samples with high confidence based merely on the softmax probability of the model output is extremely easy to optimize the model in the wrong direction, resulting in a degradation of model performance.

\subsection{Confidence Calibration}
\label{sec:2-3}
Confidence calibration is to calibrate the outputs (also known logits) of original models. Although confidence correction has been extensively studied in CV and NLP \cite{article1,article6,article19,article20,article30}, confidence correction has been rarely studied in Deep graph learning. CaGCN \cite{article40} was the first to study the problem of confidence calibration in GCNs. In CaGCN, the confidence is first corrected according to the assumption that the confidence of neighboring nodes tends to be the same and then generates pseudo labels. Unlike CaGCN, our proposed DCC-GCN model focuses on achieving a more reliable selection of high-confidence and low-confidence samples through dual-channel consistency. Then, DCC-GCN can constantly improve the model’s performance by calibrating for low-confidence samples vulnerable to misclassification.

\section{Methods}
\label{3-Methods}
The goal of our method, DCC-GCN, is to select reliable low-confidence samples and calibrate the feature embeddings of the low-confidence samples accurately. Before introducing our method, we briefly summarize the basic concepts of graph convolution in GCNs.

\subsection{Preliminaries}
\label{sec:3-0}
The GCN \cite{article8} is primarily used to process non-Euclidean graph data, generating node-level representations via message propagation. For graph $\mathcal{G}(\mathcal{V}, \mathcal{E})$, where $\mathcal{V}$ denotes the set of all nodes and $\mathcal{E}$ denotes the set of all edges, $X \in \mathbb{R}^{|\mathcal{V}| \times d}$ represents a matrix of d-dimensional input features. Feature representation of node $v \in \mathcal{V}$ at layer $k$ is obtained by aggregating its 1-hop neighbors at layer $k-1$. Use $\mathcal{N}^{1}(v)=\{u \mid(u, v) \in \mathcal{E}\}$ to denote the 1-hop neighborhood of node $v$. The graph convolution operation is defined as:

\begin{equation}
\label{equ:1}
\mathbf{h}_{v}^{l}=\sigma\left(W^{t} \cdot \text{aggregate}\left(\left\{\mathbf{h}_{u}^{l-1} \mid u \in \mathcal{N}^{1}(v)\right\}\right)\right).
\end{equation}

$\sigma(\cdot)$ represents the activation function, and aggregate usually needs to be differentiable and permutation invariant. $l$-layer GCN produces the prediction by repeating the message propagation $l$ times and then passing the final output through an MLP.

\textbf{Assumption 1}. \itshape (Symmetric Error) The GCN model $g(\cdot)$ has a classification accuracy of $p$ and makes symmetric errors for each category. We have $P[g(v)=\text{label}(v)]=p$ and $P[g(v)=k]=\frac{1-p}{c-1}$ for $\text{label}(v) \neq k \in[c]$, where $v$ is the node, and $\text{label}(v)$ is the ground-truth label of node $v$. Symmetry error is a common assumption that has been utilized in the literature \cite{article12,article13}. \upshape

\textbf{Theorem 1}. The average classification accuracy of the two GCN models is is $p_{1}$ and $p_{2}$, respectively. $N$ and $N_{r}$ represent the number of nodes in graph and the number of samples correctly classified by both GCN models respectively. The mean classification accuracy of the selected low-confidence samples, $p_{low-conf}$, is lower than the mean accuracy of the model.

\begin{equation}
\label{equ:2}
p_{low-conf}=\frac{p_{1} N-N_{r}}{\left(1-p_{1} p_{2}\right) N}<p_{1}\left(\frac{1-p_{2}}{1-p_{1} p_{2}}\right).
\end{equation}

All proofs can be found in Section A of the supplementary material.

\textbf{Analysis 1}. Calibration of the feature embedding of the low-confidence samples is the key to improving the model’s performance. As can be seen from \textbf{Theorem 1}, the classification accuracy of low-confidence samples is quite low. For example, the theoretical upper bound for $p_{low-conf}$ is 0.364 when $p_{1}=0.8$ and $p_{2}=0.7$, which is significantly lower than the average classification accuracy, limiting the overall performance of the model severely.

\textbf{Assumption 2}. \itshape The calibrated classification accuracy of the low-confidence samples is lower than the accuracy of the model, that is, $p_{low-conf}^{\prime}<p_{1}$. \upshape

\textbf{Theorem 2}. Calibrating for low confidence samples gives an upper bound to the improvement in accuracy compared to the original GCN. $\gamma$ depends on the correlation of the two models.

\begin{equation}
\label{equ:3}
p_{GAIN}<\left(1-p_{1}\right)\left[p_{1} p_{2}+\frac{\left(1-p_{1}\right)\left(1-p_{2}\right)}{c-1}+\gamma\right].
\end{equation}

All proofs can be found in Section B of the supplementary material.

\textbf{Analysis 2}. According to the \textbf{Theorem 2}, the upper bound of model improvement accuracy $p_{GAIN}$ is determined by $p_{1}$ and $p_{2}$. Furthermore, when $p_{1}$ is fixed, the $p_{GAIN}$ is greatest when $p_{2}$ is the same as $p_{1}$. See details in Section C of the supplementary material.

\subsection{Creating Diversity in GCN Models}
\label{sec:3-1}
The dual-channel uses two GCN models, each checked for the other, to filter out low-confidence samples that are difficult to classify. Diversity among two channels in DCC-GCN plays an important role in the training process, which ensures models do not converge to the same parameters.
\par According to \textbf{Analysis 2}, the maximum performance improvement bound is obtained when the accuracy of second channel model is the same as the first model. Following \cite{article9}, we construct a feature graph based on the feature matrix $X \in \mathbb{R}^{N \times M}$, which as input, can achieve comparable results to the original GCN. We first clustering the node features using \itshape KNN\upshape. When clustering node features, the higher the similarity of the representations, the higher the value of cosine similarity. We use the inverse of the cosine similarity to measure the distance between different feature representations.

\begin{equation}
\label{equ:4}
d\left(\boldsymbol{x}_{i}, \boldsymbol{x}_{j}\right)=\frac{1}{s\left(\boldsymbol{x}_{i}, \boldsymbol{x}_{j}\right)}=\frac{\left|\boldsymbol{x}_{i}\right|\left|\boldsymbol{x}_{j}\right|}{\boldsymbol{x}_{i} \cdot \boldsymbol{x}_{j}}.
\end{equation}

And then, we choose top $k$ similar node pairs for each node to set edges and finally get the adjacency matrix $A^{\prime}$. Therefore, the feature graph obtained by \itshape KNN \upshape clustering $\mathcal{G}^{\prime}=\left(A^{\prime}, X\right)$.

\begin{figure}
\centering
  \includegraphics[width=0.75\textwidth]{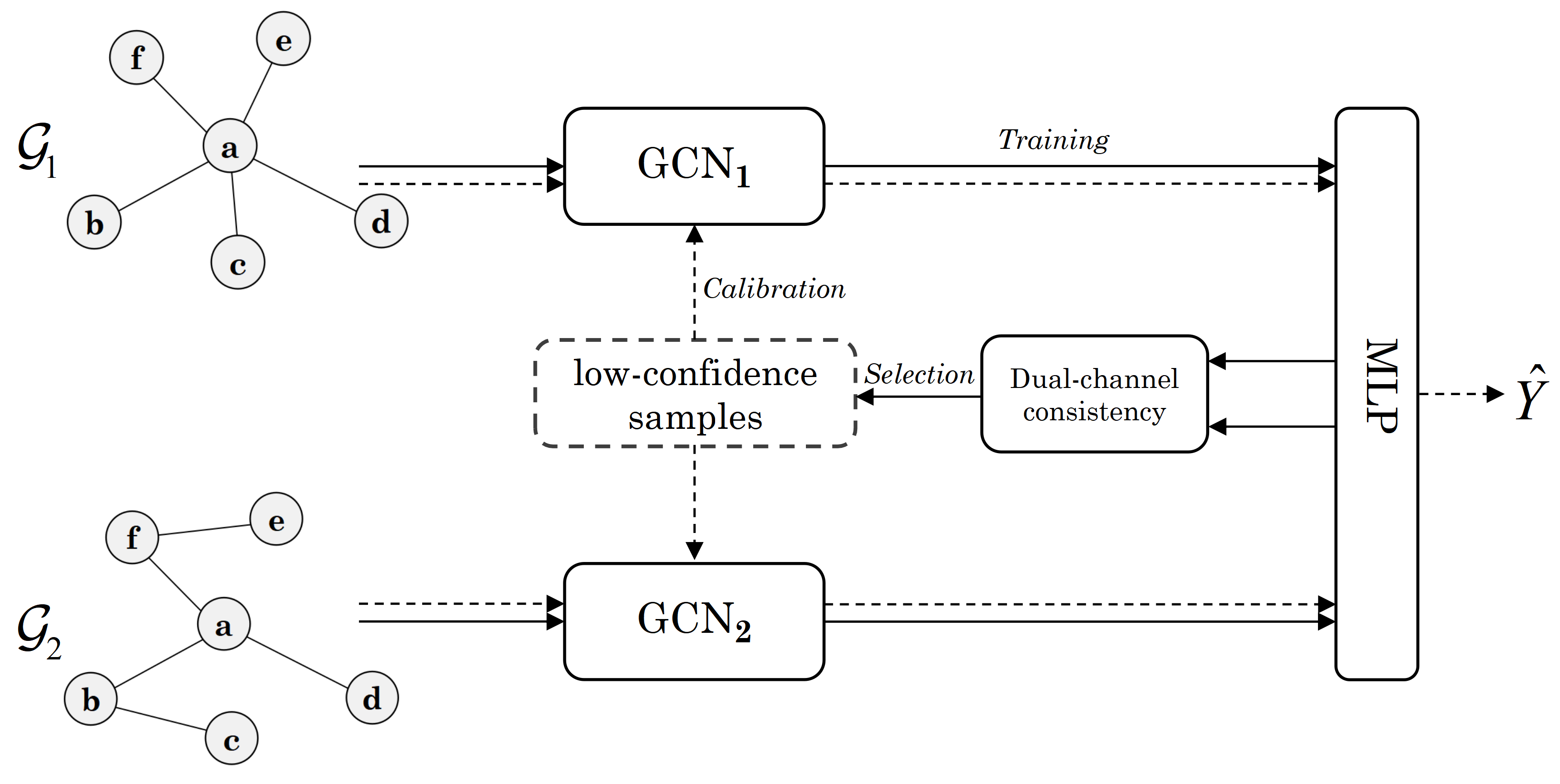}
\caption{Schematic overview of DCC-GCN. Solid lines represent the process of semi-supervised training and selection of low-confidence samples, and dashed lines represent the calibration of low-confidence samples. The inputs are the graphs $\mathcal{G}$ and $\mathcal{G}^{\prime}$, and the feature matrix $X$.}
\label{fig:2}       
\end{figure}

\subsection{Dual-Channel Consistency based GCN}
\label{sec:3-2}
Following ConfGCN \cite{article3}, DCC-GCN uses co-variance matrix based symmetric Mahalanobis distance to define the distance between two nodes. Formally, for nodes $u$ and $v$, with label distributions $\boldsymbol{\mu}_{u}$ and $\boldsymbol{\mu}_{v}$ and co-variance matrices $\boldsymbol{\Sigma}_{u}$ and $\boldsymbol{\Sigma}_{v}$, distance between them is defined as follows:

\begin{equation}
\label{equ:5}
d_{M}(u,v)=\left(\boldsymbol{\mu}_{u}-\boldsymbol{\mu}_{v}\right)^{T}\left(\boldsymbol{\Sigma}_{u}^{-1}+\boldsymbol{\Sigma}_{v}^{-1}\right)\left(\boldsymbol{\mu}_{u}-\boldsymbol{\mu}_{v}\right),
\end{equation}

where $\boldsymbol{\mu}_{v} \in \mathbb{R}^{c}$ and $\boldsymbol{\Sigma}_{v} \in \mathbb{R}^{c \times c}$.  $\boldsymbol{\mu}_{v, k}$ represents the score that node $v$ belongs to label $k$ and $\left(\boldsymbol{\Sigma}_{v}\right)_{k k}$ denotes the variance of $\boldsymbol{\mu}_{v, k}$. An important property of the distance $d_{M}(u,v)$ is that if diagonal matrices has a large eigenvalue, then the constraint that requires $\boldsymbol{\mu}_{u}$ and $\boldsymbol{\mu}_{v}$ to be close will be relaxed.
Then, we define $r_{u v}=\frac{1}{d_{M}(u, v)}$, the influence of node $u$ on node $v$ during message propagation.

\par $\hat{A}^{\prime}$ and $\hat{A}$ represent the normalized adjacency matrix of feature graph and topology graph, respectively. In the first channel, $X=[\boldsymbol{h}_{v}^{0}]$ and $A=[\alpha_{u v}]$ are used as inputs to the GCN model and the output of the $l$-th layer of node $v$ can be expressed as:

\begin{equation}
\label{equ:6}
\boldsymbol{h}_{v}^{l}=\sigma\left(\sum_{u \in \mathcal{N}_{v}} r_{u v} \alpha_{u v} \boldsymbol{h}_{u}^{l-1} \Theta^{l}\right).
\end{equation}

\subsection{Select the Low-confidence Samples}
\label{sec:3-3}
Due to overconfidence, high-confidence and low-confidence samples cannot be accurately selected using only a single model. Therefore, We determined high-confidence and low-confidence samples based on dual-channel consistency. Specifically, for the two channels, we expand its output feature matrix $\left[\boldsymbol{h}_{v}^{l}\right]$ to two times its dimensionality and input the expanded output into the MLP to obtain the classification result. The samples with the same classification result for the two GCN models in dual channels are the high-confidence samples, which form the set $\mathcal{V}_{high-conf}$. The samples with different classification results are the low-confidence samples, which form the set $\mathcal{V}_{low-conf}$.

\subsection{Calibration of Low Confidence Samples}
\label{sec:3-4}
\begin{figure}
\centering
  \includegraphics[width=0.85\textwidth]{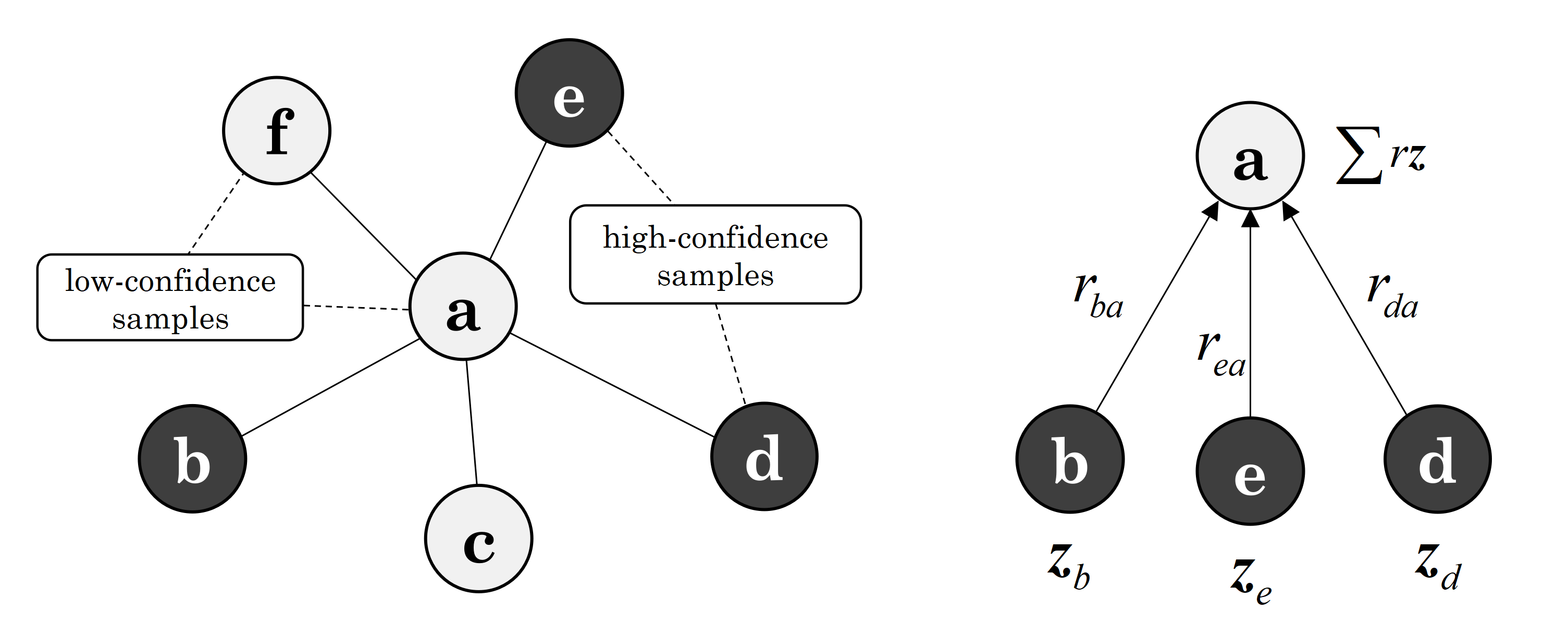}
\caption{Dark shaded nodes indicate high-confidence nodes and light grey nodes indicate low-confidence nodes. As shown in the figure on the right, the embedding of the low-confidence node a is constructed by summing the embeddings of the high-confidence samples in its 1-hop neighbors (for simplicity, $m$ is set to 1 in the figure).}
\label{fig:3}       
\end{figure}

According to \textbf{Analysis 1}, the accuracy of low-confidence samples is quite low, which restricts the performance of the model. Therefore, the calibration of low confidence samples can improve the model's performance. For high-confidence nodes, $v \in \mathcal{V}_{high-conf}$, the representations of output layer $\boldsymbol{z}_{v}= \boldsymbol{h}_{v}^{l}$. For the low-confidence samples $u \in \mathcal{V}_{low-conf}$, it is difficult to obtain the correct labels directly using the embeddings of the outputs due to the low classification accuracy of the low-confidence samples. As shown in Figure \ref{fig:3}, the embedding of the node $u$ is obtained by aggregating the embeddings of the neighbouring high-confidence nodes.

\begin{equation}
\label{equ:7}
\boldsymbol{z}_{u}=\sum_{v \in \mathcal{N}_{high-c o n f}^{m}(u)} r_{v u} \boldsymbol{h}_{v}^{l},
\end{equation}

\noindent where $\mathcal{N}_{high-conf}^{m}(u)=\mathcal{N}^{m}(u) \cap \mathcal{V}_{high-conf}$ represents the set of nodes of high-confidence in the $m$-hop neighbors of $v$. $m$ represents the hop of the neighbours used for the calibration, and in the experiments $m$ was set to 2.

The final node representations are created by concatenating the two channel representations, $\hat{\boldsymbol{z}}_{v}=contact\left(\boldsymbol{z}_{v}, \boldsymbol{z}_{v}^{\prime}\right)$, where $\boldsymbol{z}_{v}$ and $\boldsymbol{z}_{v}^{\prime}$ denote the node representations of the first and second channel, respectively. The final representations matrix could be obtained: $Z=\left[\hat{\boldsymbol{z}}_{v}\right]$. Let $V_{L}$ be the set of training nodes and $Y$ be the one-hot label matrix, and converting the model output embedding $Z$ to $\hat{Y}$ by utilizing a linear layer:

\begin{equation}
\label{equ:8}
\hat{Y}=\text{softmax}(W \cdot Z+\boldsymbol{b}),
\end{equation}

\noindent  where $\hat{Y}$ is the predicted label.

\subsection{Optimization Objective}
\label{sec:3-5}
Following \cite{article39}, we include the following two terms and cross-entropy loss of the label in objective function:

\begin{equation}
\label{equ:9}
L_{smooth}=\sum_{(u, v) \in \mathcal{E}}\left(\boldsymbol{\mu}_{u}-\boldsymbol{\mu}_{v}\right)^{T}\left(\boldsymbol{\Sigma}_{u}^{-1}
+\boldsymbol{\Sigma}_{v}^{-1}\right)\left(\boldsymbol{\mu}_{u}-\boldsymbol{\mu}_{v}\right),
\end{equation}

\begin{equation}
\label{equ:10}
L_{label}=\sum_{v \in V_{L}}\left(\boldsymbol{\mu}_{v}-\boldsymbol{Y}_{v}\right)^{T}\left(\boldsymbol{\Sigma}_{v}^{-1}+\frac{1}{\varphi} \boldsymbol{I}\right)\left(\boldsymbol{\mu}_{v}-\boldsymbol{Y}_{v}\right),
\end{equation}

where $\frac{1}{\varphi} \boldsymbol{I} \in \mathbb{R}^{L \times L}$ and ${\varphi}>0$. $L_{smooth}$ enforce neighboring nodes to be close to each other, and $L_{label}$ enforce label distribution of nodes in $V_{L}$ close to their input label distribution. Finally, we include the standard cross-entropy loss for semi-supervised multi-class classification over all the labeled nodes.

\begin{equation}
\label{equ:11}
L_{cross}=-\sum_{v \in V_{L}} Y \ln Z.
\end{equation}

Label distributions $\boldsymbol{\mu}_{v}$ and co-variance matrices $\boldsymbol{\Sigma_{v}}$ jointly with other parameters $\Theta^{l}$ is obtained by minimizing the objective function:

\begin{equation}
\label{equ:12}
L=L_{cross}+\lambda_{1} L_{smooth}+\lambda_{2} L_{label}.
\end{equation}

\section{Experiment}
\label{4-Experiment}
In this section, we conduct experiments on several datasets to demonstrate the effectiveness of our proposed DCC-GCN model.
\subsection{Datasets}
\label{sec:4-1}
We chose the four benchmark citation dataset datasets: Cora, Citeseer, Pubmed \cite{article8} and CoraFull \cite{article10}, where nodes denote documents, undirected edges denote citations between documents, and the categories of nodes are labelled according to the topic of the paper. We have also selected four other publicly available datasets. Nodes in the ACM dataset \cite{article15} denote papers, and edges indicate the presence of common authors for two papers. The above datasets all use the representation of keywords in articles as features. Flickr \cite{article18} is an image sharing website that allows users to share photos and videos. It is a social network in which users are represented by nodes and edges,and nodes are organized into nine classes according to their interests. UAI2010 is a dataset has been tested in GCN for community detection in \cite{article26}. Table \ref{tab:1} shows an overview of eight datasets.

\begin{table}[h]\small
\centering
\caption{Details of the datasets used in the paper.}
\label{tab:1}       
\begin{tabular}{ccccc}
\hline\noalign{\smallskip}
Dataset	&Classes &Features &Nodes &Edges \\
\noalign{\smallskip}\hline\noalign{\smallskip}
Cora         &7  &1,433  &2,708  &5,429  \\
Citeseer	 &6  &3,703  &3,327  &4,732  \\
Pubmed	     &3  &500    &19,717 &44,338  \\
CoraFull	 &70 &8,710  &19,793 &65,331 \\
ACM	         &3  &1,870  &3,025	 &13,128\\
Flickr       &9  &12,047 &7,575  &239,738\\
UAI2010      &19 &4,973	 &3,067  &28,311\\

\noalign{\smallskip}\hline
\end{tabular}
\end{table}

\subsection{Baseline}
\label{sec:4-2}
To evaluate our proposed DCC-GCN, we compared it to the following baseline:
\begin{itemize}
\item
\textbf{LP} \cite{article14} Propagation of a node's label to neighboring nodes based on the distance between the node and the neighboring nodes.
\end{itemize}

\begin{itemize}
\item
\textbf{GCN} \cite{article8} is spectral graph convolution localized first-order approximation used for semi-supervised learning on graph-structured data.
\end{itemize}

\begin{itemize}
\item
\textbf{GAT} \cite{article11} uses the attention mechanism to determine the weights of node neighbors. By adaptively assigning weights to different neighbors, the performance of the graph neural network is improved.
\end{itemize}

\begin{itemize}
\item
\textbf{DGCN} \cite{article22} uses two parallel simple feedforward networks, the difference is only when the input graph structure information is different, and the parameters of the two parallel graph convolution modules are shared.
\end{itemize}

\begin{itemize}
\item
\textbf{AGNN} \cite{article23} introduces attention mechanism in the propagation layer, so that the attention of the central node to the neighbor node will be different in the process of feature aggregation.
\end{itemize}

\begin{itemize}
\item
\textbf{ConfGCN} \cite{article3} jointly estimate label scores and their confidence values in GCN, using these estimated confidence values to determine the effect of one node on another.
\end{itemize}

\begin{itemize}
\item
\textbf{E-GCN} \cite{article4} uses both the given label and the estimated label for the topology optimization of the GCN model.
\end{itemize}
\subsection{Node Classification Results}
\label{sec:4-3}
In order to make a fair comparison, we do not set the validation set and use all samples except the training set as the test set. We use Accuracy (ACC) and macro F1-score (F1) to evaluate performance of models. 20 samples from each category were selected as the training set, and in Table \ref{tab:2} we present the average precision of all results over 10 runs.

\begin{table}[h]
\centering
\caption{Classification accuracies of compared methods on CoraFull, ACM, Flickr, and UAI2010 datasets. We ignore 3 classes with less than 50 nodes in CoraFull dataset (since the test set has too few nodes in these classes).}
\label{tab:2}       
\resizebox{\textwidth}{!}{
\begin{tabular}{cccccccccc}
\hline\noalign{\smallskip}
Dataset	&Metrics &LP &GCN &GAT &AGNN &DGCN &ConfGCN &E-GCN &DCC-GCN\\
\noalign{\smallskip}\hline\noalign{\smallskip}
\multirow{2}*{CoraFull}&ACC&30.2&54.3&55.4&55.6&54.9&\underline{55.9}&55.7&\textbf{56.8}\\
	                   &F1 &27.4&49.7&51.6&52.0&50.7&\underline{52.3}&52.1&\textbf{52.3}\\
\noalign{\smallskip}\hline\noalign{\smallskip}
\multirow{2}*{ACM}     &ACC&61.5&86.5&86.2&85.5&85.9&87.9&\underline{88.1}&\textbf{88.9}\\
	                   &F1 &61.1&86.4&86.0&83.2&84.3&85.3&\underline{85.6}&\textbf{87.8}\\
\noalign{\smallskip}\hline\noalign{\smallskip}
\multirow{2}*{Flickr}  &ACC&24.3&41.5&38.5&38.8&42.1&\underline{44.1}&43.8&\textbf{72.0}\\
	                   &F1 &20.8&40.2&36.8&37.1&40.1&\underline{43.3}&42.7&\textbf{72.8}\\
\noalign{\smallskip}\hline\noalign{\smallskip}
\multirow{2}*{UAI2010} &ACC&41.7&47.6&49.3&50.2&50.4&\underline{56.2}&55.8&\textbf{69.7}\\
                       &F1 &32.6&35.9&37.2&39.0&41.2&\underline{43.2}&42.7&\textbf{48.2}\\
\noalign{\smallskip}\hline
\end{tabular}}
\end{table}

We observe a significant improvement in the effectiveness of DCC-GCN compared to GCN on the Flickr and UAI2010 datasets. This is because our method extracts the feature information of the nodes more effectively. See Section \ref{sec:5-4} for a detailed analysis of this phenomenon.

\subsection{Results under Scarce Labeled Training Data}
\label{sec:4-4}
In order to evaluate our model more comprehensively, we set up training sets with different label rates on three datasets, Cora, Citeseer and Pubmed. For Cora and Citeseer: \{0.5\%, 1\%, 1.5\%, 2\%\} and for Pubmed: \{0.03\%, 0.05\%, 0.07\%, 0.1\%\}. The results of the semi-supervised node classification for the different models are summarised in Tables \ref{tab:3} to \ref{tab:5}.

\begin{table}[ht]\small
\centering
\caption{Classification Accuracy on Cora.}
\label{tab:3}       
\begin{tabular}{ccccc}
\multicolumn{5}{c}{\textbf{Cora Dateset}}\\
\hline\noalign{\smallskip}
\textbf{Label Rate} & 0.5\% & 1\% & 1.5\%  & 2\% \\
\noalign{\smallskip}\hline\noalign{\smallskip}
\textbf{LP}         &57.6 &61.2 &62.4 &63.4\\
\textbf{GCN}	    &54.2 &61.0 &66.2 &72.8\\
\textbf{GAT}        &54.3 &60.3 &66.5 &72.5\\
\textbf{AGNN}	    &54.8 &60.9 &66.7 &72.8\\
\textbf{DGCN}	    &56.3 &62.6 &67.3 &73.1\\
\textbf{ConfGCN}	&60.1 &63.6 &69.5 &73.5\\
\textbf{E-GCN}	    &60.8 &65.2 &70.4 &73.4\\
\noalign{\smallskip}\hline\noalign{\smallskip}
\textbf{DCC-GCN}	&\textbf{63.9} &\textbf{67.2} &\textbf{71.8} &\textbf{74.6}\\
\textbf{GAIN}	    &\textbf{3.1}  &\textbf{2.0}  &\textbf{1.4}  &\textbf{1.1}\\
\noalign{\smallskip}\hline
\end{tabular}
\end{table}

\begin{table}[ht]\small
\centering
\caption{Classification Accuracy on Citeseer.}
\label{tab:4}       
\begin{tabular}{ccccc}
\multicolumn{5}{c}{\textbf{Citeseer Dateset}}\\
\hline\noalign{\smallskip}
\textbf{Label Rate} & 0.5\% & 1\% & 1.5\%  & 2\% \\
\noalign{\smallskip}\hline\noalign{\smallskip}
\textbf{LP}	       &39.6 &43.2 &45.5 &48.2\\
\textbf{GCN}	   &46.6 &56.3 &59.8 &64.8\\
\textbf{GAT}	   &46.7 &56.6 &60.3 &65.2\\
\textbf{AGNN}	   &47.8 &57.1 &60.5 &65.3\\
\textbf{DGCN}	   &48.2 &58.3 &61.8 &65.8\\
\textbf{ConfGCN}   &49.2 &61.3 &63.2 &66.7\\
\textbf{E-GCN}	   &51.2 &60.7 &64.0 &66.4\\
\noalign{\smallskip}\hline\noalign{\smallskip}
\textbf{DCC-GCN}   &\textbf{53.3} &\textbf{62.8} &\textbf{65.3} &\textbf{67.9}\\
\textbf{GAIN}	   &\textbf{2.1}  &\textbf{1.5}  &\textbf{1.3}	 &\textbf{1.2}\\
\noalign{\smallskip}\hline
\end{tabular}
\end{table}

\begin{table}[ht]\small
\centering
\caption{Classification Accuracy on Pubmed.}
\label{tab:5}       
\begin{tabular}{ccccc}
\multicolumn{5}{c}{\textbf{Pubmed Dateset}}\\
\hline\noalign{\smallskip}
\textbf{Label Rate} &0.03\% &0.05\% &0.07\%  &0.1\% \\
\noalign{\smallskip}\hline\noalign{\smallskip}
\textbf{LP}	       &59.4 &61.8 &62.3 &63.6\\
\textbf{GCN}	   &57.2 &59.8 &63.4 &67.6\\
\textbf{GAT}	   &57.8 &60.4 &64.5 &67.8\\
\textbf{AGNN} 	   &57.9 &60.3 &64.3 &68.0\\
\textbf{DGCN} 	   &58.3 &61.0 &64.6 &68.2\\
\textbf{ConfGCN}   &60.8 &62.3 &66.1 &69.2\\
\textbf{E-GCN}	   &61.4 &62.8 &66.9 &68.9\\
\noalign{\smallskip}\hline\noalign{\smallskip}
\textbf{DCC-GCN}   &\textbf{63.3} &\textbf{63.9} &\textbf{67.8} &\textbf{70.0}\\
\textbf{GAIN}	   &\textbf{1.9}  &\textbf{1.1}  &\textbf{0.9}  &\textbf{0.8}\\
\noalign{\smallskip}\hline
\end{tabular}
\end{table}

We have the following observations:
Compared to the baseline, we proposed DCC-GCN consistently outperforms GCN, GAT, AGNN, DGCN, ConfGCN and E-GCN on all the datasets. It is evident that when labeled data is insufficient, GCN's performance suffers significantly due to inefficient label information propagation. For example, On the Cora and Pubmed datasets, GCN performs even worse than LP \cite{article14} when the training size is limited. Within a certain range, the performance improvement of DCC-GCN compared to the GCN model, demonstrating the effectiveness of this framework on graphs with few labeled nodes.

\section{Discussion}
\label{5-Discussion}
\subsection{Effect of Low-confidence Samples Calibration}
\label{sec:5-1}
\begin{figure}[h]
\centering
\subfloat[Cora]{
\begin{minipage}[h]{0.32\linewidth}
\centering
\includegraphics[width=2in]{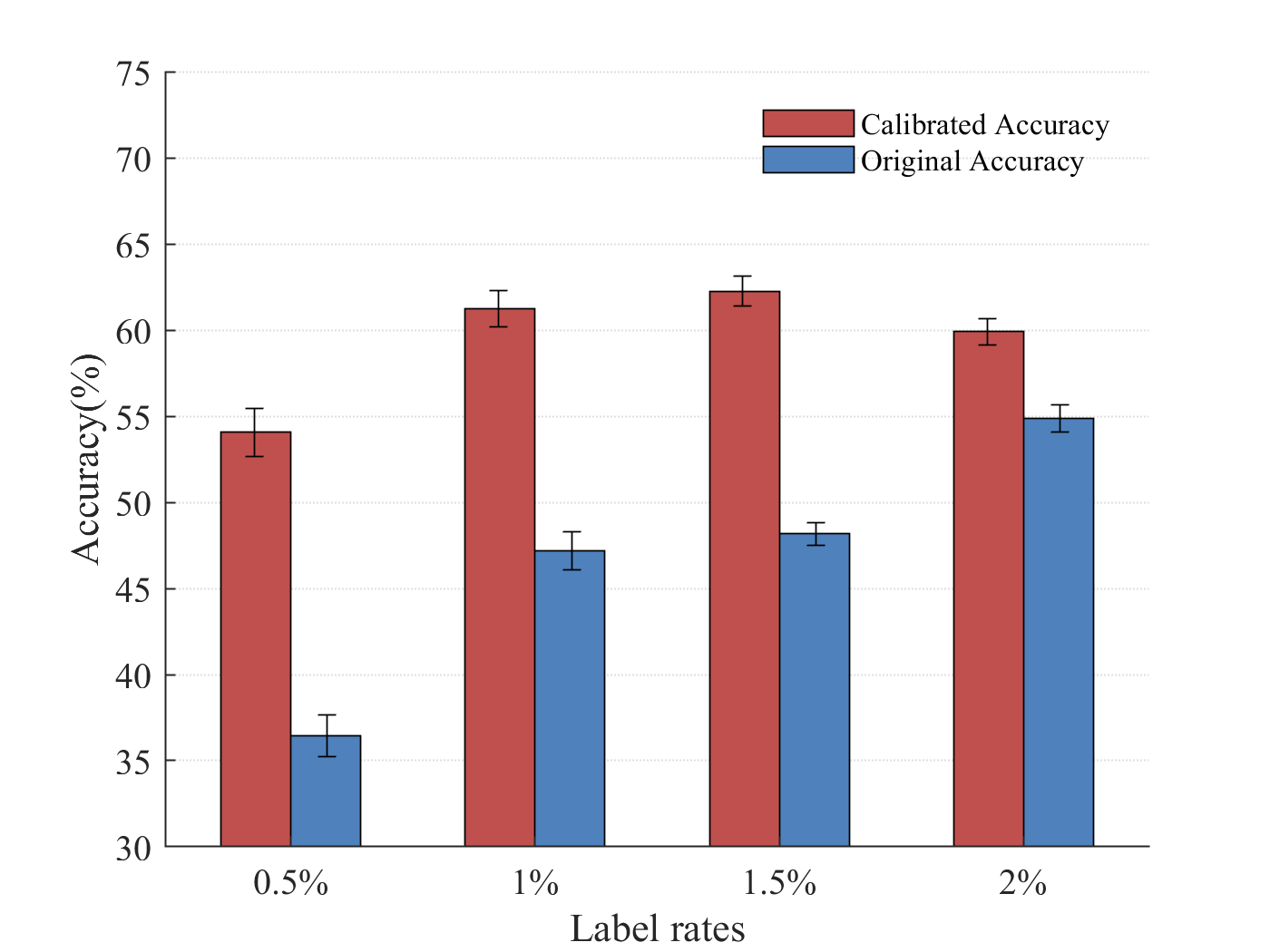}
\end{minipage}%
}%
\subfloat[Citeseer]{
\begin{minipage}[h]{0.32\linewidth}
\centering
\includegraphics[width=2in]{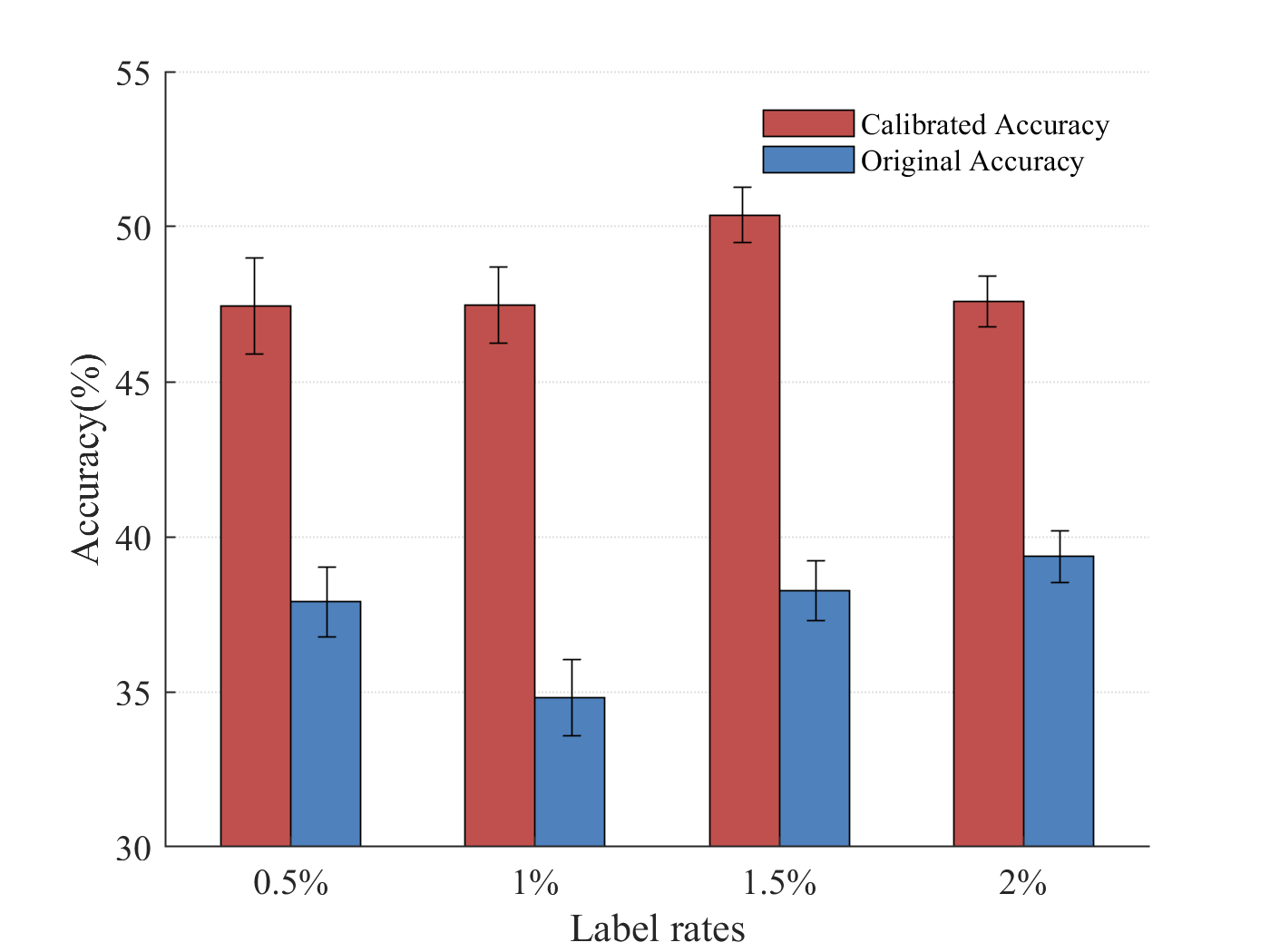}
\end{minipage}%
}%
\subfloat [Pubmed]{
\begin{minipage}[h]{0.32\linewidth}
\centering
\includegraphics[width=2in]{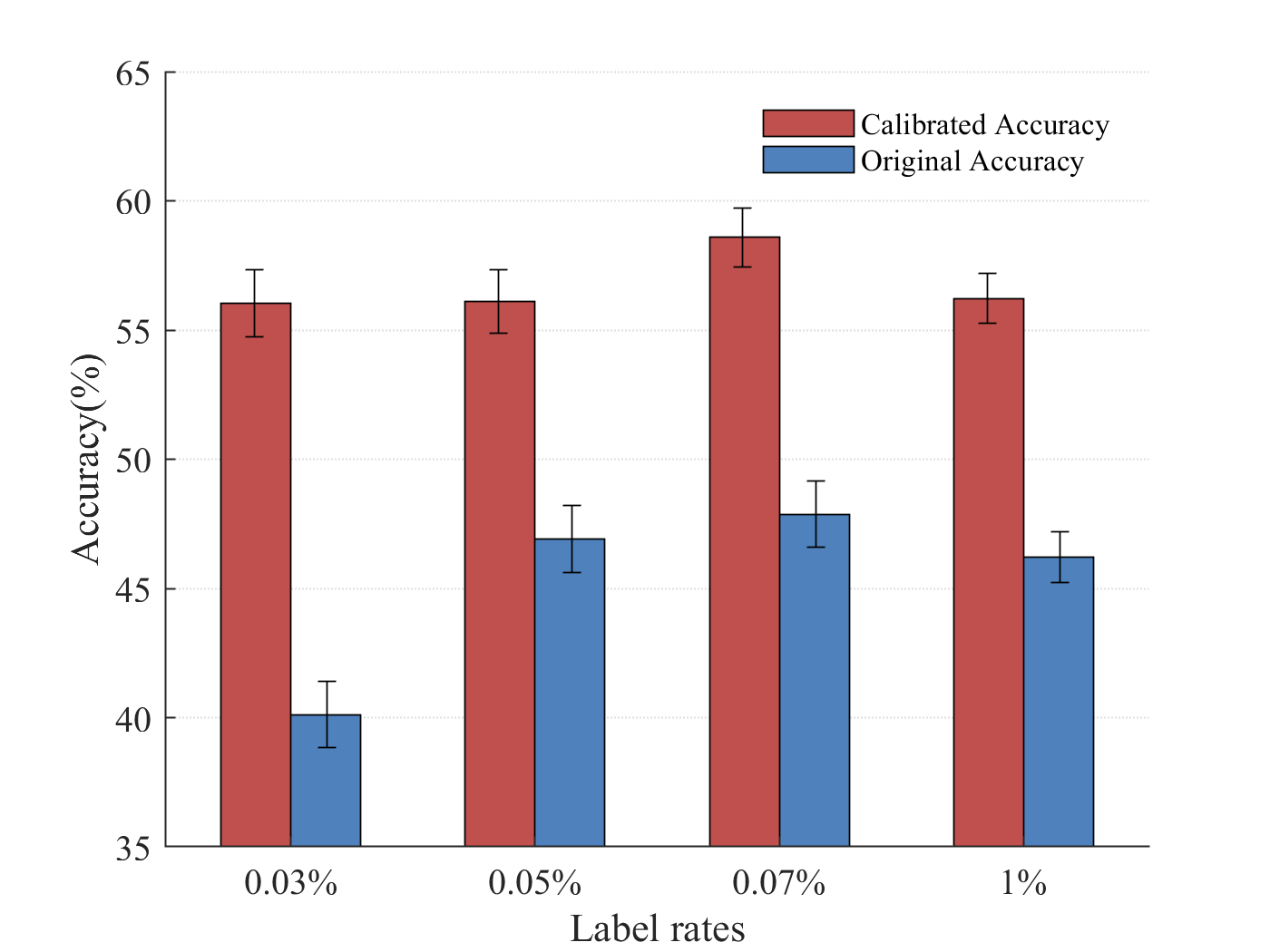}
\end{minipage}%
}%
\caption{The original and calibrated accuracy of the low-confidence samples.}
\label{fig:4}
\centering
\end{figure}

According to Analysis 1 in Section \ref{3-Methods}, the critical factor in increasing the model's performance is the Calibration of the low-confidence samples. In this section, we test the accuracy of low-confidence samples before ands after the calibration by graph structure.
Before calibrating the low-confidence samples, we first employ GCN as the base models to learn a label distribution for each node. Training epochs is taken as 100, 100, 200 for Cora, Citeseer and Pubmed under different label rates, respectively. In Figure \ref{fig:4}, the original and calibrated accuracy of the low-confidence samples are drawn separately for Cora, Citeseer, and Pubmed under different label rates. It is obvious that the accuracy of low-confidence samples after the calibration improved significantly.

\subsection{Reliability of High-confidence Samples}
\label{sec:5-2}
In SSL, training bias \cite{article31} is a common problem. When the model generates false predictions with high confidence, these false predictions will further strengthen the deviation of the model and lead to the deterioration of the model performance. Selecting more reliable high-confidence samples is the key to improve model performance.
\par In this section, we conduct a simple experiment to prove the reliability of high-confidence samples selected based on dual-channel consistency. Firstly, Obtain high-confidence samples based on DCC-GCN, and then we randomly select 100 samples to join the training of the original GCN model as pseudo-labeled samples. We also used the traditional method of obtaining pseudo-labels \cite{article42} as a comparison. The loss function of the model is:

\begin{equation}
\label{equ:13}
Loss=L_{s}+\alpha \cdot L_{p},
\end{equation}

where $\alpha=0.3$ when the number of training epochs is greater than 100, and $\alpha=0$, otherwise. $L_{s}$ and $L_{p}$ denote the cross-entropy loss for the labelled and pseudo-labelled samples, respectively. We use the same hyper-parameters and training set, test set for all datasets as in \cite{article8}. The experimental results are shown in Table \ref{tab:6}.

\begin{table}[htb]
\begin{center}
\caption{Classification Accuracy on Cora, Citeseer and Pubmed.}
\label{tab:6}
\begin{tabular}{cccc}
\hline  Method &Cora &Citeseer &Pubmed\\
\hline GCN                                        &81.5 &70.3 &79.0 \\
GCN + pseudo-labels (\cite{article42})     &81.5 &70.3 &79.1 \\
GCN + pseudo-labels (DCC-GCN)              &\textbf{83.1} &\textbf{72.1} &\textbf{79.1} \\
\hline
\end{tabular}
\end{center}
\end{table}

Experimental results show that the selection of pseudo-labels using the output of a single GCN was ineffective in improving the model's performance since they are most likely to be simple samples that do not help much in training. Adding high-confidence samples based on DCC-GCN directly into the training set as pseudo-label samples can improve the model's performance, proving that high-confidence samples can be selected reliably based on DCC-GCN.

\subsection{Analysis of Neighborhood Hop for Calibration}
\label{sec:5-3}
\begin{figure}[h]
\centering
\subfloat[Cora]{
\begin{minipage}[h]{0.32\linewidth}
\centering
\includegraphics[width=2in]{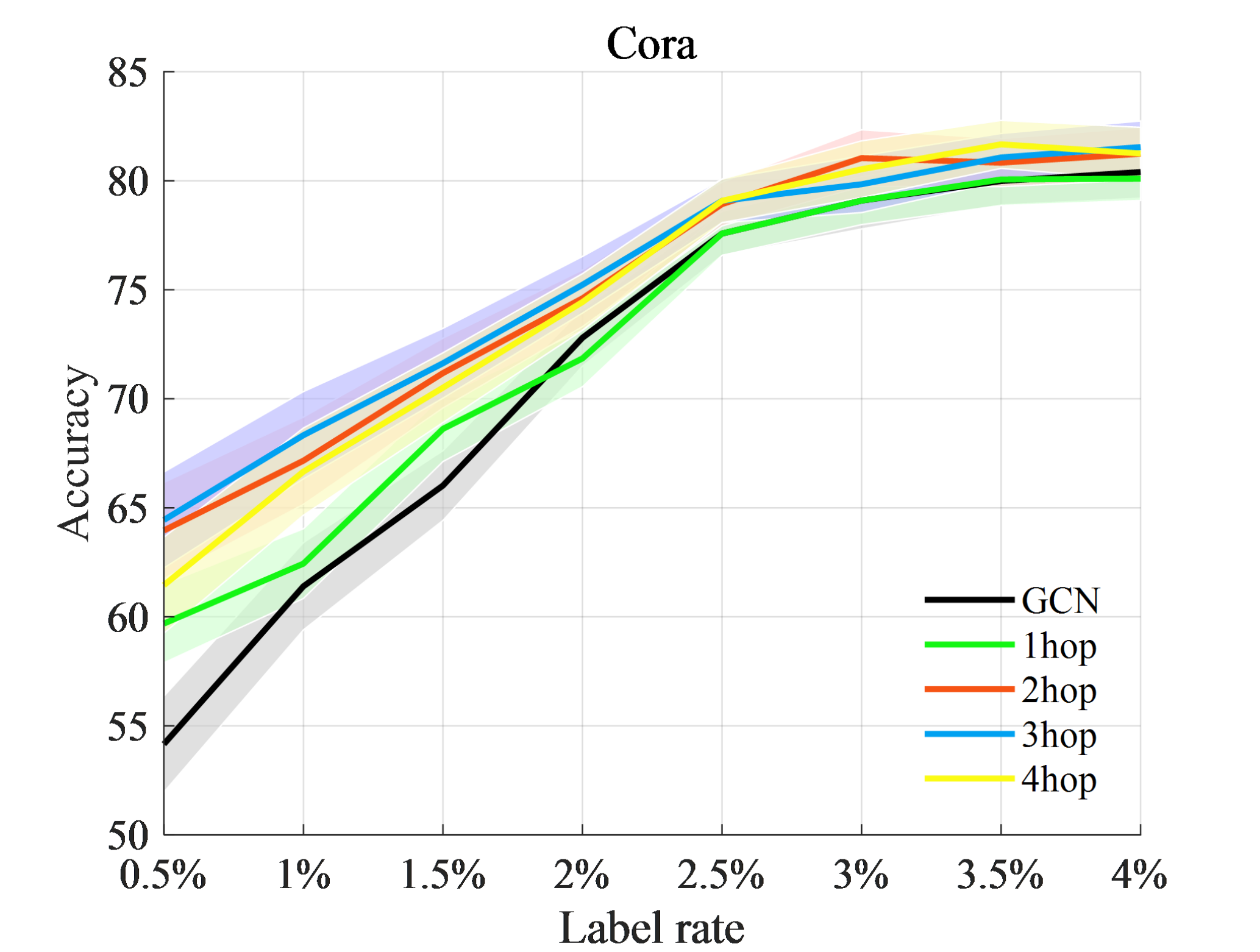}
\end{minipage}%
}%
\subfloat[Citeseer]{
\begin{minipage}[h]{0.32\linewidth}
\centering
\includegraphics[width=2in]{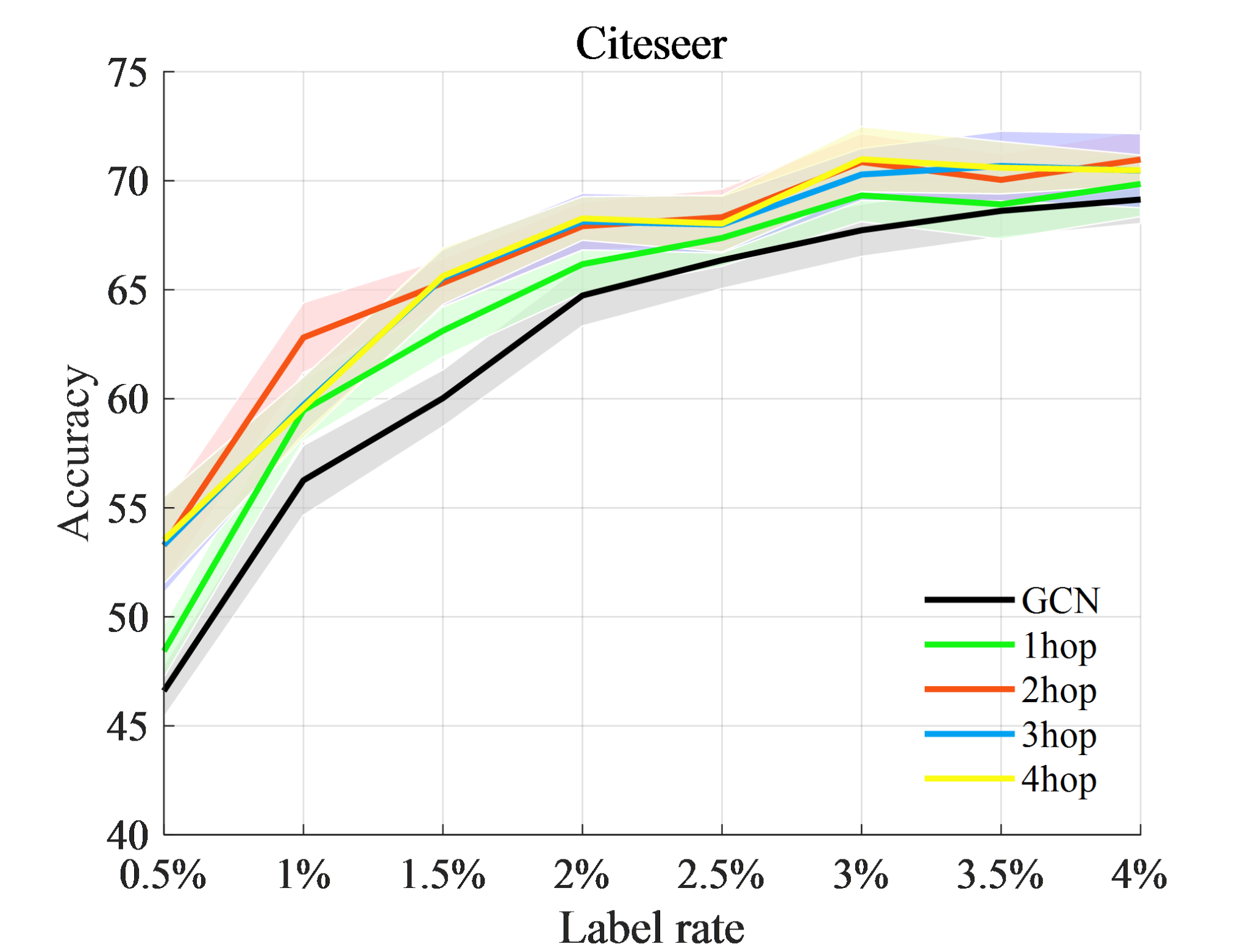}
\end{minipage}%
}%
\subfloat [Pubmed]{
\begin{minipage}[h]{0.32\linewidth}
\centering
\includegraphics[width=2in]{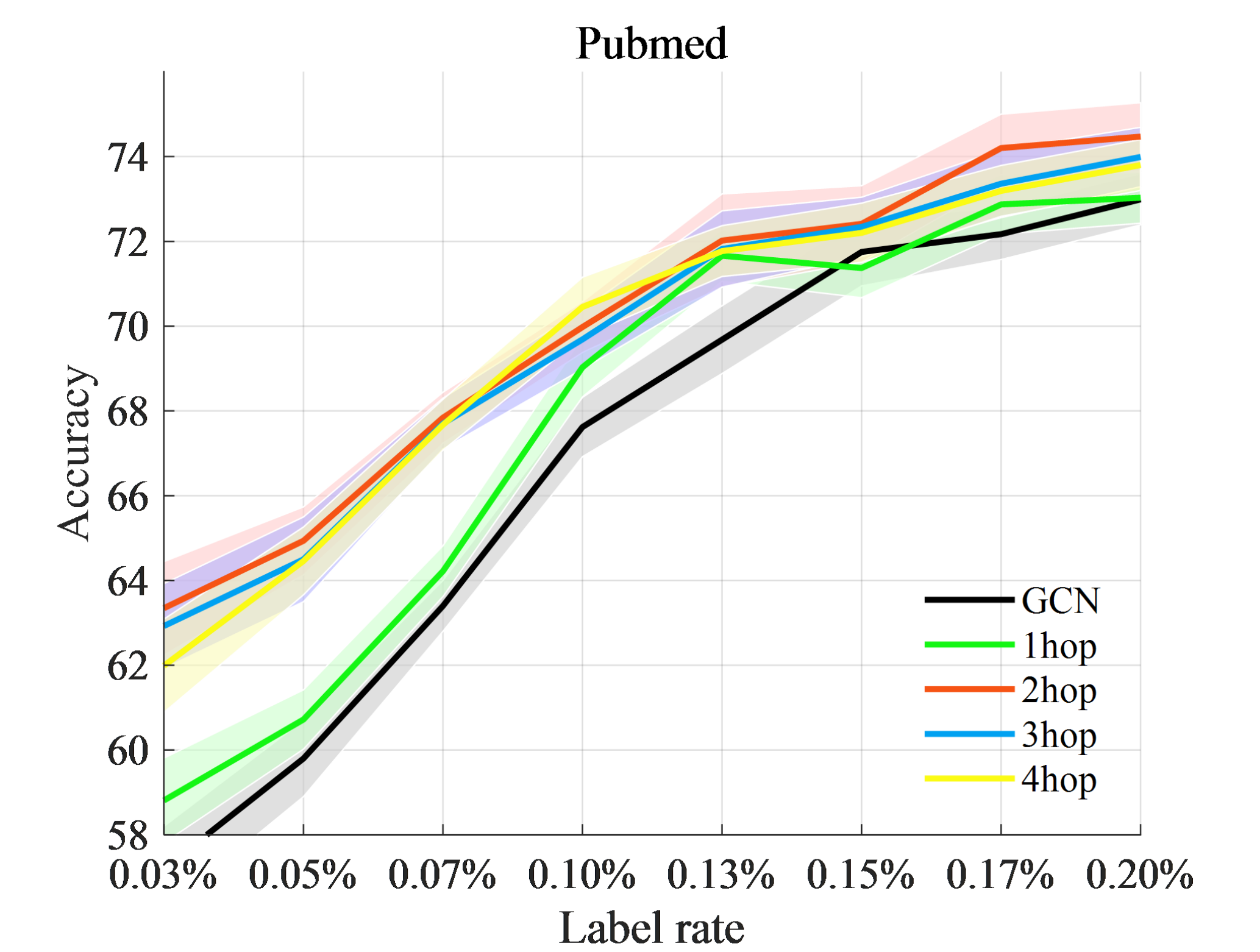}
\end{minipage}%
}%
\caption{Analysis of neighbourhood hop used during the calibration process.}
\label{fig:5}
\centering
\end{figure}

In this section, we explore the effect of the neighbourhood hop for calibration. On the Cora, Citeseer, and Pubmed datasets, we set $m$ from 1 to 4 at different labelling rates, respectively, and the experimental results are shown in Figure \ref{fig:5}.

With only the 1-hop neighbors during the calibration, DCC-GCN could not improve the accuracy significantly, and in some cases, the classification accuracy is even lower than that of GCN on the Cora and Pubmed datasets. However, it is easy to observe that DCC-GCN with hop larger than 1 for low-confidence samples calibration all outperform the original GCNs with a large margin, especially when the graph has low label rate.

\subsection{Ablation Study}
\label{sec:5-4}
In this section, we remove various parts of our model to study the impact of each component. We first demonstrate that the calibration of low confidence samples based on dual-channel consistency, has a large effect on the results compared to model without calibration, then we show that the aggregation of dual-channel is more beneficial than using one channel only.
For simplicity, we adopt DCC-GCN (w/o Calibration) and DCC-GCN (w/o Aggregation) to represent the reduced models by removing the calibration of low-confidence samples and the aggregation of dual-channel , respectively. The comparison is shown in Table \ref{tab:7}.

\begin{table}[h]\small
\centering
\caption{Ablation study of the calibration of low-confidence samples and the aggregation of dual-channel on Cora, Citeseer, and Pubmed datasets.}
\label{tab:7}       
\begin{tabular}{cccccc}
\hline\noalign{\smallskip}
Method &Metrics &Cora-Full &ACM  &Flickr &UAI2010\\
\noalign{\smallskip}\hline\noalign{\smallskip}
 \multirow{2}*{DCC-GCN} &ACC&\textbf{57.1}&\textbf{88.9}&\textbf{72.0}&\textbf{69.7}\\
	                   &F1 &\textbf{52.4}&\textbf{87.8}&\textbf{72.8}&\textbf{48.2}\\
\noalign{\smallskip}\hline\noalign{\smallskip}
\multirow{2}*{DCC-GCN(w/o Calibration)}	&ACC&55.4&86.7&70.1&65.8\\
                       &F1 &50.6&86.4&70.6&43.9\\
\noalign{\smallskip}\hline\noalign{\smallskip}
\multirow{2}*{DCC-GCN(w/o Aggregation)} &ACC&56.8&88.1&45.1&48.5\\
                       &F1&51.1 &87.2&43.2&36.3\\
\noalign{\smallskip}\hline
\end{tabular}
\end{table}

The results of the original model are consistently better than all the other two variants, indicating the effectiveness of using the two constraints together. Apparently, ACC and F1 decrease when the aforementioned component is dropped from the framework. It reveals that our proposed calibration of low-confidence samples and aggregation of dual-channel modules are able to improve the performance on semi-supervised learning task substantially.

\begin{figure}[h]
\centering
\subfloat[GCN]{
\begin{minipage}[h]{0.32\linewidth}
\centering
\includegraphics[width=2in]{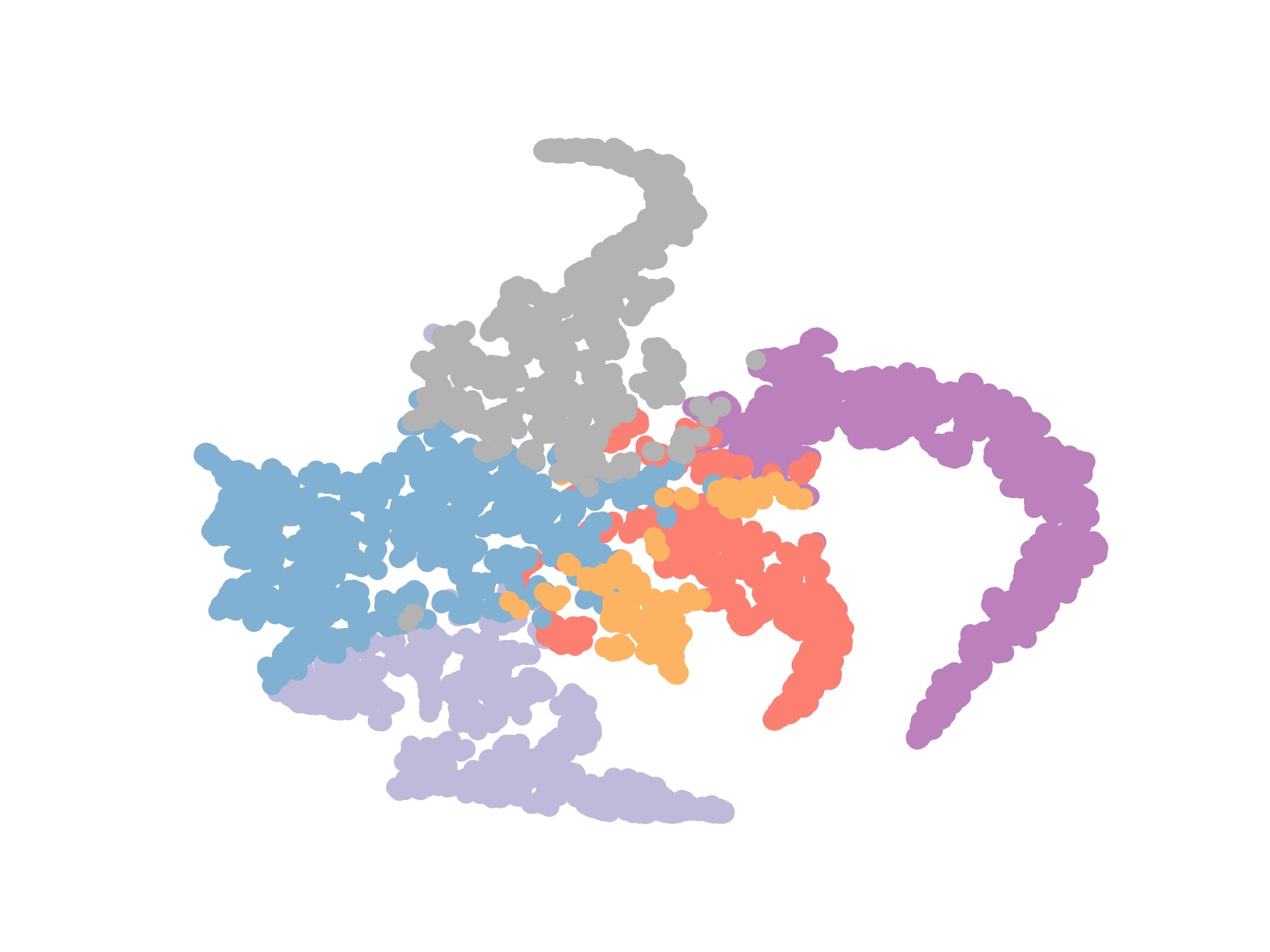}
\end{minipage}%
}%
\subfloat[MLP]{
\begin{minipage}[h]{0.32\linewidth}
\centering
\includegraphics[width=2in]{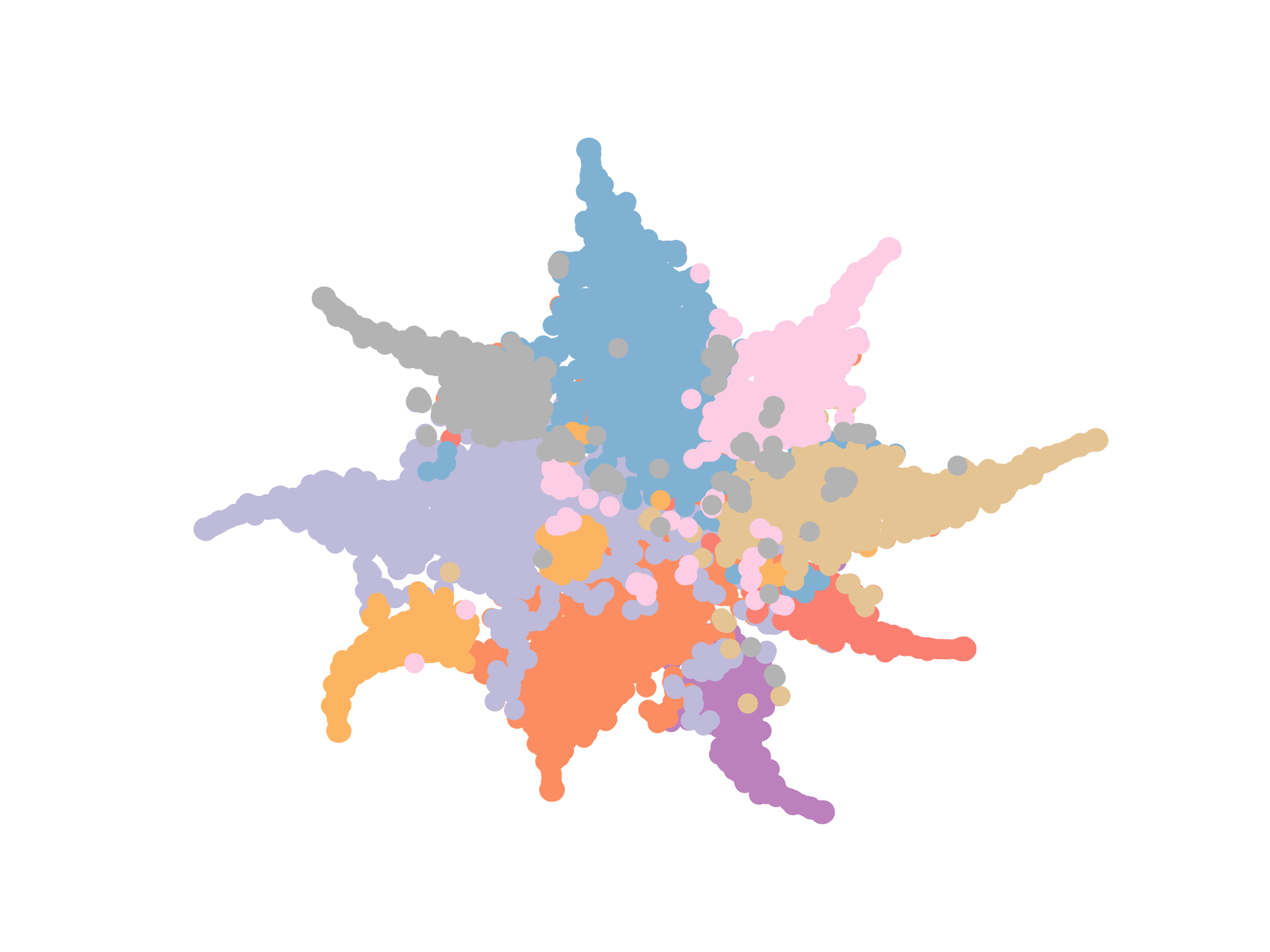}
\end{minipage}%
}%
\subfloat [DCC-GCN]{
\begin{minipage}[h]{0.32\linewidth}
\centering
\includegraphics[width=2in]{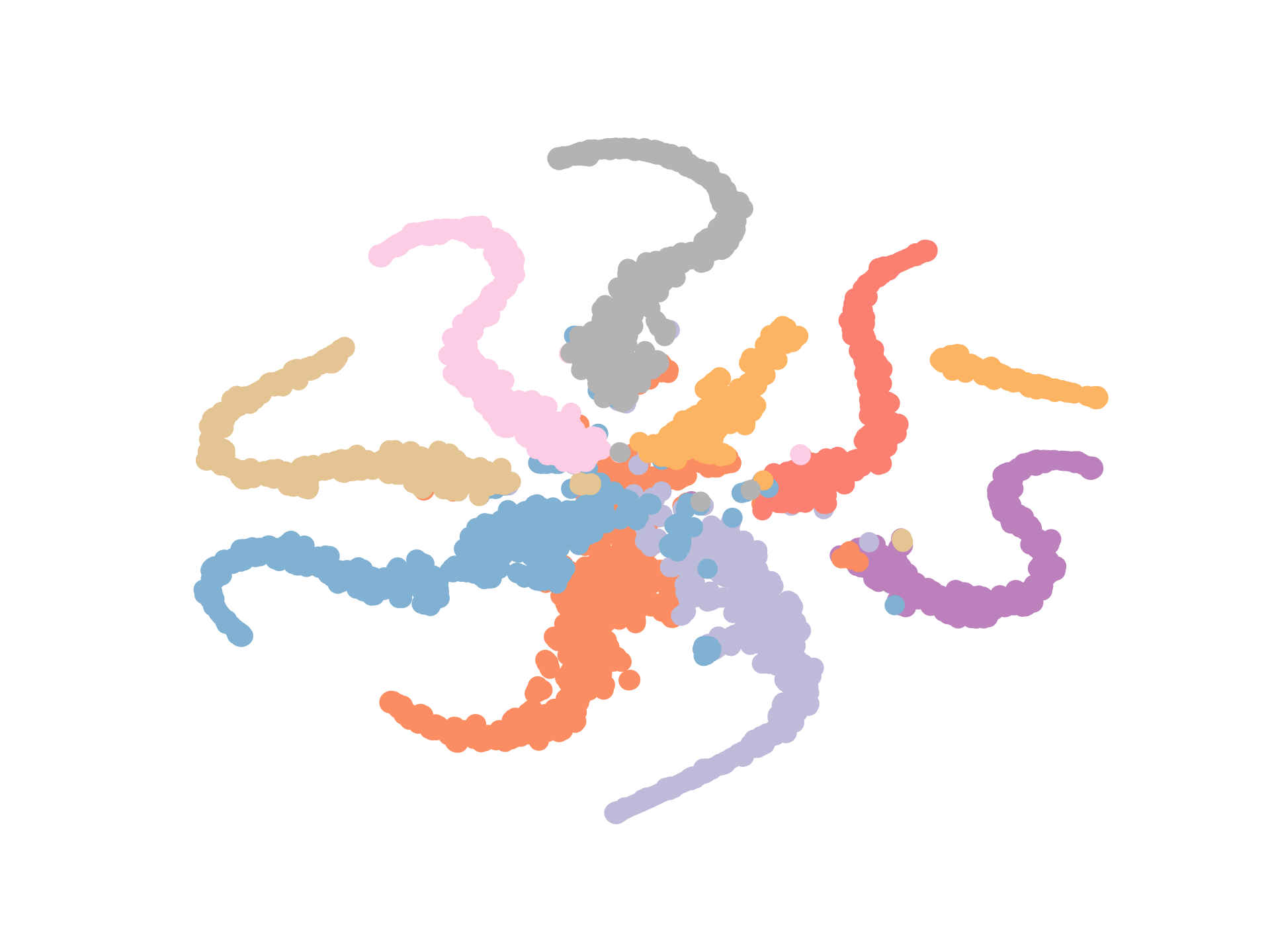}
\end{minipage}%
}%
\caption{T-SNE embeddings of nodes in Flickr dataset. (a), (b) and (c) denote representations learned by GCN, MLP, and DCC-GCN respectively.}
\label{fig:6}
\centering
\end{figure}

In addition, the improvement of DCC-GCN over DCC-GCN (w/o Calibration) is more substantial on the UAI2010 and Flickr. To reveal this, we visualize the embedding results of Flickr dataset generated by GCN, KNN and DCC-GCN via using t-SNE \cite{article41} in Figure \ref{fig:6}.

The classification accuracy of GCN and MLP is 41.5\% and 57.3\%, respectively, which indicates that the features of nodes contain more information than the graph topology structure. As can be observed, the embeddings generated by DCC-GCN can exhibit more coherent clusters compared with the other two methods. This is due to DCC-GCN can adaptively learn information between topology and node features, training two classifiers based on the graph $\mathcal{G}$ and $\mathcal{G}^{\prime}$ in dual-channel.

\section{Conclusion}
\label{6-Conclusion}
In this paper, we propose a semi-supervised learning framework for GCN named DCC-GCN based on dual-channel consistency. DCC-GCN could select low-confidence samples through dual-channel consistency and calibrate them based on samples with high-confidence in the neighbourhood. DCC-GCN can learn the topology of the graph that is most suitable for the task of graph-based semi-supervised node classification. Experiments on several benchmarks demonstrate that our method is superior to state-of-the-art graph-based semi-supervised learning methods.

\section{Acknowledgements}
\label{7-Acknowledgements}
This work was supported by the National Key R\&D Program of China under Grant No. 2017YFB1002502.

\bibliographystyle{elsarticle-harv}
\bibliography{DCCGCNref}

\clearpage
\section*{Appendix}
\appendix

\subsection{A. Theorem 1 in Section III}
\label{sec:8-1}
\textbf{Proof}.
The inputs for the two GCN models are graphs $\mathcal{G}$ and $\mathcal{G}^{\prime}$, respectively, and the average accuracy of classification is $p_{1}$ and $p_{2}$, respectively. There are $N$ nodes in graph $\mathcal{G}$ and $\mathcal{G}^{\prime}$. The number of samples with the same classification result for both GCN models $N_{a}$ is the number of samples $N_{r}$ correctly classified by both GCN models added to the number of samples $N_{w}$ incorrectly classified as the same class by the two models.
\par If the two GCN models are completely independent:

\begin{equation}
\label{equ:13}
N_{r}=p_{1} p_{2} N, N_{e}=C_{c-1}\left(\frac{1-p_{1}}{c-1}\right)\left(\frac{1-p_{2}}{c-1}\right) N=\frac{\left(1-p_{1}\right)\left(1-p_{2}\right)}{c-1} N,
\end{equation}

\begin{equation}
\label{equ:14}
N_{a}=N_{r}+N_{w}=\left[p_{1} p_{2}+\frac{\left(1-p_{1}\right)\left(1-p_{2}\right)}{c-1}\right] N.
\end{equation}

In fact, due to the correlation between the two GCN models, the number of samples correctly classified by the two models $N_{r}=p_{1} p_{2} N+\alpha N$, and the number of samples incorrectly classified by the two models as the same category $N_{w}=\frac{\left(1-p_{1}\right)\left(1-p_{2}\right)}{c-1} N+\beta N$. We can get:

\begin{equation}
\label{equ:15}
N_{a}=N_{r}+N_{w}=\left[p_{1} p_{2}+\frac{\left(1-p_{1}\right)\left(1-p_{2}\right)}{c-1}+\alpha+\beta\right] N.
\end{equation}

The more significant the correlation between the two GCN models, the more excellent $\alpha$ and $\beta$ will be,and $\alpha>0$, $\beta>0$. For simplicity, let $\gamma=\alpha+\beta$. The average classification accuracy for low-confidence samples is $p_{low-conf}$. For the first GCN model, the number of correctly classified samples can be expressed as $p_{1}N$. The number of correctly classified samples can also be expressed as:

\begin{equation}
\label{equ:16}
N_{a}+\left[1-p_{1} p_{2}-\frac{\left(1-p_{1}\right)\left(1-p_{2}\right)}{c-1}-\gamma\right] N \cdot p_{low-conf}.
\end{equation}

Equation can be obtained as:

\begin{equation}
\label{equ:17}
N_{a}+\left[1-p_{1} p_{2}-\frac{\left(1-p_{1}\right)\left(1-p_{2}\right)}{c-1}-\gamma\right] N \cdot p_{low-conf}=p_{1} N.
\end{equation}

An upper bound for $p_{low-conf}$ can be obtained:

\begin{equation}
\label{equ:18}
p_{low-conf}=\frac{p_{1}-p_{1} p_{2}-\frac{\left(1-p_{1}\right)\left(1-p_{2}\right)}{c-1}-\gamma}{1-p_{1} p_{2}-\frac{\left(1-p_{1}\right)\left(1-p_{2}\right)}{c-1}-\gamma}<p_{1}\left(\frac{1-p_{2}}{1-p_{1} p_{2}}\right).
\end{equation}

\subsection{B. Theorem 2 in Section III}
\label{sec:8-2}
\textbf{Proof}. The average classification accuracy of the low-confidence samples is $p_{low-conf}$, the average classification accuracy after the calibration is $p_{low-conf}^{\prime}$, and the performance improvement of the model is $p_{GAIN}$. After the low-confidence samples calibration, the Equation \ref{equ:17} can be rewritten as:

\begin{equation}
\label{equ:19}
N_{a}+\left[1-p_{1} p_{2}-\frac{\left(1-p_{1}\right)\left(1-p_{2}\right)}{c-1}-\gamma\right] N \cdot p_{low-conf}^{\prime}=\left(p_{1}+p_{GAIN}\right)N.
\end{equation}

Substitute $N_{a}=\left[p_{1} p_{2}+\frac{\left(1-p_{1}\right)\left(1-p_{2}\right)}{c-1}+\gamma\right]N$ into the above expression:

\begin{equation}
\label{equ:20}
\left[p_{1} p_{2}+\frac{\left(1-p_{1}\right)\left(1-p_{2}\right)}{c-1}+\gamma\right]+\left[1-p_{1} p_{2}-\frac{\left(1-p_{1}\right)\left(1-p_{2}\right)}{c-1}-\gamma\right] p_{\text {low-conf}}^{\prime}=p_{1}+p_{GAIN}.
\end{equation}

Since $p_{low-conf}^{\prime}<p_{1}$, we can obtain the inequality:

\begin{equation}
\label{equ:21}
p_{1}+p_{G A I N}<\left[p_{1} p_{2}+\frac{\left(1-p_{1}\right)\left(1-p_{2}\right)}{c-1}+\gamma\right]+\left[1-p_{1} p_{2}-\frac{\left(1-p_{1}\right)\left(1-p_{2}\right)}{c-1}-\gamma\right] p_{1}.
\end{equation}

Simplify inequality,

\begin{equation}
\label{equ:22}
p_{GAIN}<\left(1-p_{1}\right)\left[p_{1} p_{2}+\frac{\left(1-p_{1}\right)\left(1-p_{2}\right)}{c-1}+\gamma\right].
\end{equation}

\subsection{C. Analysis 2 in Section III}
\label{sec:8-3}

\begin{figure}
\centering
  \includegraphics[width=0.95\textwidth]{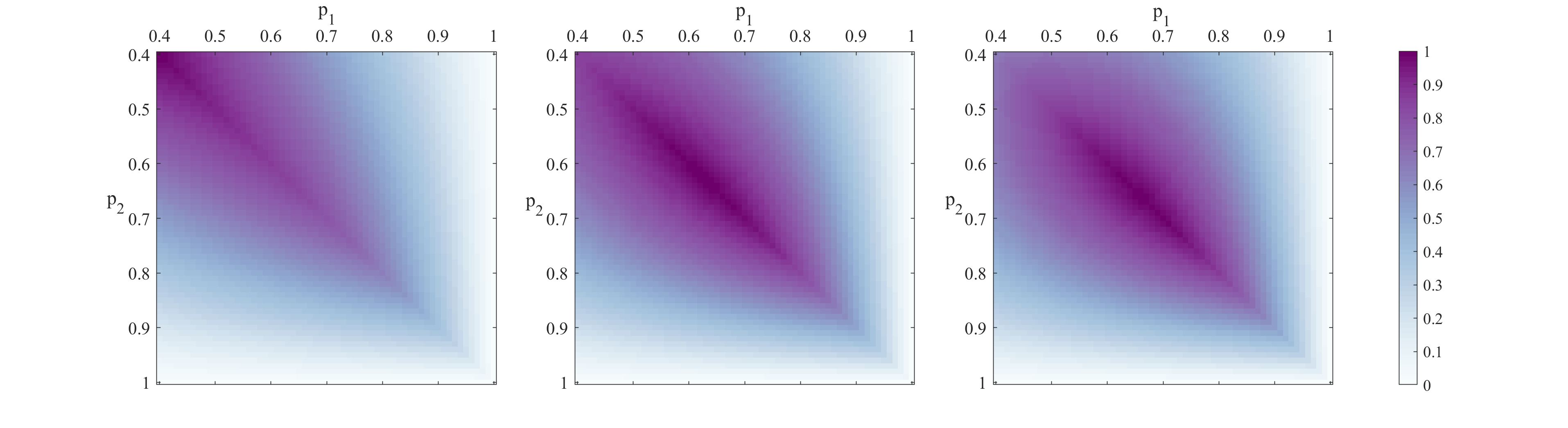}
\caption{Variation of the upper bound of $p_{GAIN}$ with $p_{1}$, $p_{2}$ ($c$ is 3, 7 and 70 respectively).}
\label{fig:7}       
\end{figure}

$\gamma$ is related to the correlation between the dual-channel models and is determined by the model and its parameters and data set. According to Assumption 2, $p_{low-conf}^{\prime}<p_{1}$. Similarly, we can obtain: $p_{low-conf}^{\prime}<p_{2}$. The upper bound of model improvement accuracy is determined by $p_{1}$ and $p_{2}$.
For simplicity, let $\gamma=0$, using $p_{1}$ and $p_{2}$ as the X and Y axes respectively, draw the upper bound of the accuracy of the promotion when the category $c$ is 3, 7 and 70 in Figure \ref{fig:7} respectively.

We can observe that the bigger the difference between $p_{1}$ and $p_{2}$, the lower the upper bound of $p_{GAIN}$. When the value of $p_{1}$ is fixed and the value of $p_{2}$ is the same as $p_{1}$, the upper bound of $p_{GAIN}$ is the highest.

\subsection{D. Implementation Details}
\label{sec:8-4}

In this paper, all models use a 2-layer GCN with ReLU as the activation function. We train the model for a fixed number of epochs, specifically. 200, 200, 500, 1000 epochs for Cora, Citeseer, Pubmed and CoraFull, respectively, 300 for acm, 200 for Flickr, and 200 for UAI2010. All models along with $\boldsymbol{\mu}$ are initialized with Xavier initialization \cite{article24}, and matrix $\boldsymbol{\Sigma}$ is initialized with identity. All models were trained using Adam SGD optimizer \cite{article25} on all datasets. Our models were implemented in PyTorch version 1.6.0. When constructing feature graph, $k\in\{2 ...10\}$ for k-nearest neighbor graph. All dataset-specific hyper-parameters are summarized in Table \ref{tab:8}.

\begin{table}[h]
\centering
\caption{Hype-parameter specifications.}
\label{tab:8}       
\resizebox{\textwidth}{!}{
\begin{tabular}{ccccccccc}
\hline\noalign{\smallskip}
Dataset &\tabincell{c}{Learning\\rate} &\tabincell{c}{Weight\\decay} &\tabincell{c}{Hidden\\dimension1} &\tabincell{c}{Hidden\\dimension2} &Dropout &$k$ &$\lambda_{1}$ &$\lambda_{2}$\\
\noalign{\smallskip}\hline\noalign{\smallskip}
Cora        &5e-3&1e-5&256&128&0.5&6 &0.25&0.5\\
Citeseer    &1e-3&1e-5&768&128&0.5&6 &0.25&0.5\\
Pubmed      &1e-3&1e-5&768&256&0.5&3 &0.25&0.5\\
CoraFull    &2e-4&1e-5&512&128&0.5&10&0.25&0.5\\
ACM         &1e-4&5e-4&768&256&0.5&9 &0.2&0.8\\
Flickr      &1e-4&5e-4&512&128&0.5&5 &0.4&0.8\\
UAI2010     &1e-4&5e-4&512&128&0.5&6 &0.35&0.7\\
\noalign{\smallskip}\hline
\end{tabular}}
\end{table}





\end{document}